\pdfoutput=1

\documentclass[11pt]{article}
\usepackage{graphicx}
\usepackage{authblk}
\usepackage[final]{acl}

\usepackage{times}
\usepackage{latexsym}
\usepackage{amsmath}
\usepackage{amssymb} 
\usepackage{booktabs}
\usepackage{tabularx}
\usepackage{mathtools}
\usepackage{arydshln} 
\usepackage{multirow} 
\usepackage{stfloats}
\usepackage[figuresright]{rotating}
\usepackage{float}
\usepackage{url}
\usepackage[T1]{fontenc}
\usepackage[utf8]{inputenc}
\usepackage[russian,english]{babel}
\makeatletter
\newcommand\email[2][]%
   {\newaffiltrue\let\AB@blk@and\AB@pand
      \if\relax#1\relax\def\AB@note{\AB@thenote}\else\def\AB@note{\relax}%
        \setcounter{Maxaffil}{0}\fi
      \begingroup
        \let\protect\@unexpandable@protect
        \def\thanks{\protect\thanks}\def\footnote{\protect\footnote}%
        \@temptokena=\expandafter{\AB@authors}%
        {\def\\{\protect\\\protect\Affilfont}\xdef\AB@temp{#2}}%
         \xdef\AB@authors{\the\@temptokena\AB@las\AB@au@str
         \protect\\[\affilsep]\protect\Affilfont\AB@temp}%
         \gdef\AB@las{}\gdef\AB@au@str{}%
        {\def\\{, \ignorespaces}\xdef\AB@temp{#2}}%
        \@temptokena=\expandafter{\AB@affillist}%
        \xdef\AB@affillist{\the\@temptokena \AB@affilsep
          \AB@affilnote{}\protect\Affilfont\AB@temp}%
      \endgroup
       \let\AB@affilsep\AB@affilsepx
}
\makeatother
\usepackage[T1]{fontenc}

\usepackage[utf8]{inputenc}

\usepackage{microtype}

\usepackage{inconsolata}
\title{Unveiling the Power of Source: Source-based Minimum Bayes Risk Decoding for Neural Machine Translation}

\author[1]{Boxuan Lyu}
\author[2]{Hidetaka Kamigaito}
\author[1]{Kotaro Funakoshi}
\author[1]{Manabu Okumura}
\affil[1]{Institute of Science Tokyo}
\affil[2]{Nara Institute of Science and Technology}
\email{\url{{lyu,funakoshi,oku}@lr.first.iir.isct.ac.jp}}
\email{\url{kamigaito.h@is.naist.jp}}

\def\code#1{\texttt{#1}}
\begin{document}
\maketitle
\begin{abstract}
Maximum a posteriori decoding, a commonly used method for neural machine translation (NMT), aims to maximize the estimated posterior probability. However, high estimated probability does not always lead to high translation quality. Minimum Bayes Risk (MBR) decoding (\citealp{kumar2004minimum}) offers an alternative by seeking hypotheses with the highest expected utility.

Inspired by Quality Estimation (QE) reranking which uses the QE model as a ranker (\citealp{fernandes-etal-2022-quality}), we propose source-based MBR (sMBR) decoding, a novel approach that utilizes quasi-sources (generated via paraphrasing or back-translation) as ``support hypotheses'' and a reference-free quality estimation metric as the utility function, marking the first work to solely use sources in MBR decoding.
Experiments show that sMBR outperforms QE reranking and the standard MBR decoding.
Our findings suggest that sMBR is a promising approach for NMT decoding.\footnote{Our models and code are available at: \url{https://github.com/vlaks425/sMBR}}
\end{abstract}

\section{Introduction}

Neural Machine Translation (NMT) models typically aim to select a hypothesis with the highest estimated posterior probability during decoding, an approach known as Maximum A Posteriori (MAP) decoding. Beam search (\citealp{DBLP:journals/corr/abs-1211-3711}; \citealp{sutskever2014sequence}), which balances computational cost and search accuracy, has become the standard approximate decoding method for MAP.

\begin{figure}
    \centering
    \includegraphics[width=1.0\linewidth]{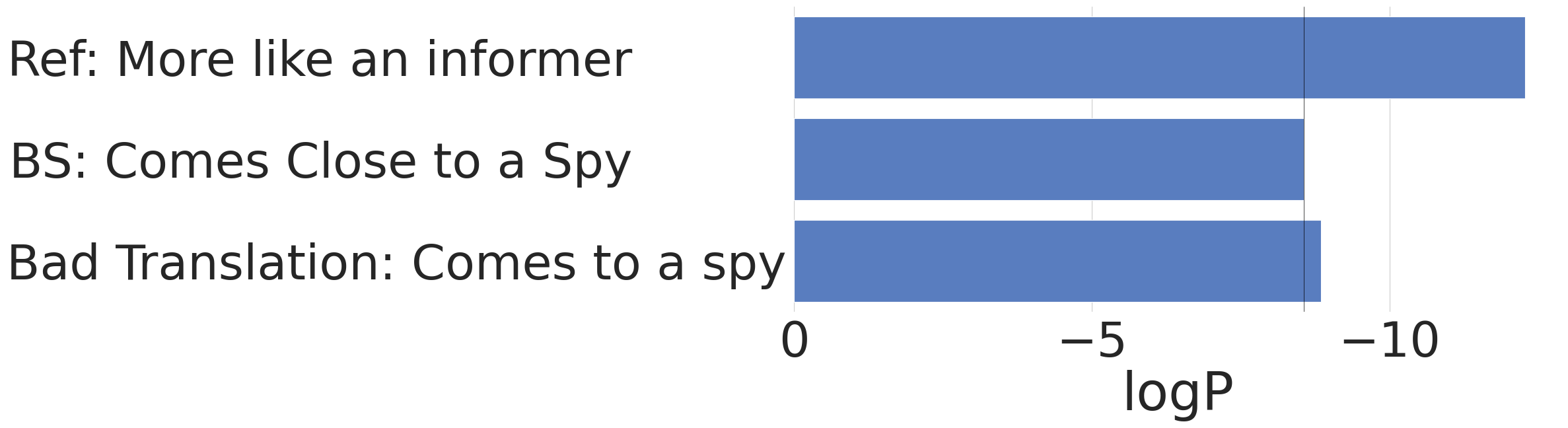}
    \caption{Example of De$\rightarrow$En, with source \textit{``Kommt einem Spitzel nahe''}. BS denotes beam search. The estimated log probability of a human reference is lower than that of the beam search output, and even lower than that of a bad translation.}
    \label{fig:logP}
\end{figure}

However, the underlying assumption of beam search - that estimated probability is a good proxy for translation quality - has been challenged by evidence showing that estimated probability and quality do not always correlate positively (\citealp{ott2018analyzing}; \citealp{freitag-etal-2021-experts}). For example, Fig~\ref{fig:logP} illustrates a case where a human reference translation has a lower estimated probability than the hypothesis generated by beam search, and even lower than that of a poor translation. Furthermore, the true MAP output is sometimes an empty string or overly brief translation (\citealp{koehn-knowles-2017-six}; \citealp{murray-chiang-2018-correcting}; \citealp{ott2018analyzing}; \citealp{stahlberg-byrne-2019-nmt}). These suggest that solely searching for high estimated probability hypotheses may not be an effective strategy for improving quality.

Given the limitations of using estimated probability as a proxy for quality, an attractive alternative is to directly target translation quality during decoding (\citealp{freitag-etal-2022-high}). Minimum Bayes Risk (MBR) decoding, proposed in the era of statistical machine translation (\citealp{kumar2004minimum}; \citealp{tromble-etal-2008-lattice}), aims to find the hypothesis with the highest expected utility with respect to a set of hypotheses called ``support hypotheses''. Traditionally, surface-based evaluation metrics like BLEU (\citealp{papineni-etal-2002-bleu}) were used as the utility function in MBR decoding (\citealp{eikema-aziz-2020-map}; \citealp{eikema-aziz-2022-sampling}). However, these metrics have shown limited correlation with human judgments (\citealp{mathur-etal-2020-tangled}; \citealp{freitag-etal-2023-results}), hindering the widespread adoption of MBR decoding based on them. Recent work has explored using state-of-the-art neural metrics, such as COMET (\citealp{rei-etal-2022-comet}), as utility functions for MBR decoding (\citealp{freitag-etal-2022-high}; \citealp{fernandes-etal-2022-quality}; \citealp{freitag-etal-2023-epsilon}), showing promising improvements in human evaluations.

Moreover, advances in reference-free evaluation metrics (\citealp{rei-etal-2021-references}; \citealp{rei-etal-2022-cometkiwi}; \citealp{rei-etal-2023-scaling}) have enabled their direct application to hypothesis reranking, which we refer to as Quality Estimation (QE) reranking (\citealp{fernandes-etal-2022-quality}). QE reranking selects the hypothesis with the highest reference-free quality estimation score among the candidate hypotheses. However, QE reranking remains understudied compared to MBR decoding.

Inspired by QE reranking, which uses the QE model as a reranker,
we propose a novel approach called \textbf{source-based MBR} (sMBR) decoding, which uses quasi-sources generated by paraphrasing or back-translation as ``support hypotheses'' and a QE metric as the utility function. 
This marks the first work to solely use sources as support hypotheses in MBR decoding, breaking the long-standing tradition of relying on using other hypotheses to approximate true utility for this purpose. 
See Fig~\ref{fig:overview} for a overview of our methodology.
We provide empirical evidence through comprehensive experiments on three translation directions in both classic (large transformer models trained from scratch, including high-resource and low-resource sub-setups) and large language model (LLM) setups, demonstrating that sMBR outperforms QE reranking and the standard MBR decoding.
These findings suggest that sMBR is a promising NMT decoding approach.

\section{Decoding Methods in NMT}
Decoding can be viewed as two phases: \textit{hypothesis generation} and \textit{decision}. Specifically, in the hypothesis generation phase, a certain generation method, such as beam search, is used to generate ${N}$ hypotheses from the model ${\{h_0,h_1,\ldots,h_{N-1}\}}$. Then, in the decision phase, ${N}$ decision scores need to be computed for each of these ${N}$ hypotheses ${\{{score}_{0},{score}_{1},\ldots,{score}_{{N-1}}\}}$. Finally, the hypothesis with the highest decision score is selected as the final output.
\subsection{MAP decoding}
Given 
a source sentence ${x}$
and a hypothesis space ${\mathcal{H}}$, the translation model ${P_{mt}(\cdot\mid x)}$ estimates the probability of any hypothesis $h \in \mathcal{H}$. MAP decoding aims to find the hypothesis ${y^{\mathit{MAP}}}$ that maximizes the probability:
\begin{align}
y^{\mathit{MAP}}  = \mathop{\mathrm{argmax}}\limits_{h \in \mathcal{H}}{P_{mt}}
(h\mid x).
\end{align}

In other words, MAP decoding simply takes the estimated probability as the decision score.

However, considering all possible hypotheses in ${\mathcal{H}}$ is computationally intractable. Therefore, beam search is widely used as an efficient approximation of MAP decoding, balancing the trade-off between computational cost and search accuracy.

Increasing the beam size leads to searching for hypotheses with higher estimated probabilities. However, in practice, when the beam size exceeds 5 or 10, it often leads to a performance degradation instead (\citealp{DBLP:conf/aaai/TuLSLL17}; \citealp{koehn-knowles-2017-six}). This phenomenon is known as the \textit{beam search curse}, considered one of the six challenges of NMT (\citealp{koehn-knowles-2017-six}).
\subsection{MBR decoding}
\label{sec:mbr}
Unlike MAP decoding, which aims to find the highest estimated probability hypothesis, MBR decoding seeks the hypothesis that minimizes the expected loss (or equivalently, maximizes the expected utility) with respect to a set of hypotheses, called ``support hypotheses''.

In practice, it is common to use a set of hypotheses from a model as support hypotheses. Formally, let ${\mathcal{S} \subseteq {\mathcal{H}}}$ be a set of support hypotheses from model $P_{mt}(\cdot \mid x)$, for support hypothesis ${h_s \in \mathcal{S}}$, 
MBR decoding selects the hypothesis ${y^{\mathit{MBR}}}$ that has the least risk:
\begin{align}\hspace{-2.3mm}
y^{\mathit{MBR}} & =  \mathop{\mathrm{argmin}}\limits_{h \in \mathcal{H}} {\mathbb{E}_{h_s \in \mathcal{S}}\left[L(h_s, h) \mid x\right]}
\\ & =  \mathop{\mathrm{argmin}}\limits_{h \in \mathcal{H}}
\sum_{h_s \in \mathcal{S}} L(h_s, h) {P_{mt}(h_s\mid x)}.
\end{align}

In practice, it is common to use a utility function, correlated to human evaluation results, such as BLEU or COMET, as an alternative to the loss function $L(\cdot,\cdot)$. Thus, the purpose of MBR decoding is actually to select the hypothesis with the maximum expected utility. 
In addition, the hypothesis space is usually too large 
to traverse all the hypotheses to find a translation that satisfies the above conditions. Therefore, a set of hypotheses ${\mathcal{C}}$ from the hypothesis space ${\mathcal{H}}$, called ``candidate hypotheses'', is often used as a representative of the whole hypothesis space.
Combining these two points, for a given utility function ${u(\cdot,\cdot)}$, the MBR decoding objective can be reformulated as:

\begin{align}
\hspace{-2.5mm}
y^{\mathit{MBR}} \approx \mathop{\mathrm{argmax}}\limits_{h \in \mathcal{C}}
\sum_{h_s \in \mathcal{S}} u(h_s, h) {P_{mt}}(h_s\mid x).
\label{def:mbr}
\end{align}

A widely used practice is 
to use the same set of hypotheses for both $\mathcal{C}$ and $\mathcal{S}$, and
to assume that each $h_s$ have the same probability, instead of the estimated probability given by the model. 
That is, the expected utility for a chosen hypothesis is approximated by averaging its utilities to other hypotheses. Hypotheses can be obtained through, for example, beam search or ancestral sampling\footnote{In the context of NMT, ancestral sampling refers to sampling from the entire vocabulary without any pruning.}.
Formally, the objective of MBR can be approximated as:
\begin{align}
y^{\mathit{MBR}} &\approx \mathop{\mathrm{argmax}}\limits_{h \in \mathcal{C}} 
score^{\mathit{MBR}}_{h},\\
score^{\mathit{MBR}}_{h} & = \frac{1}{|\mathcal{S}|} \sum_{h_s \in \mathcal{S}} u(h_s, h).
\label{def:MBR_appro}
\end{align}


\begin{figure*}[t]
    \centering
    \includegraphics[width=1.0\linewidth]{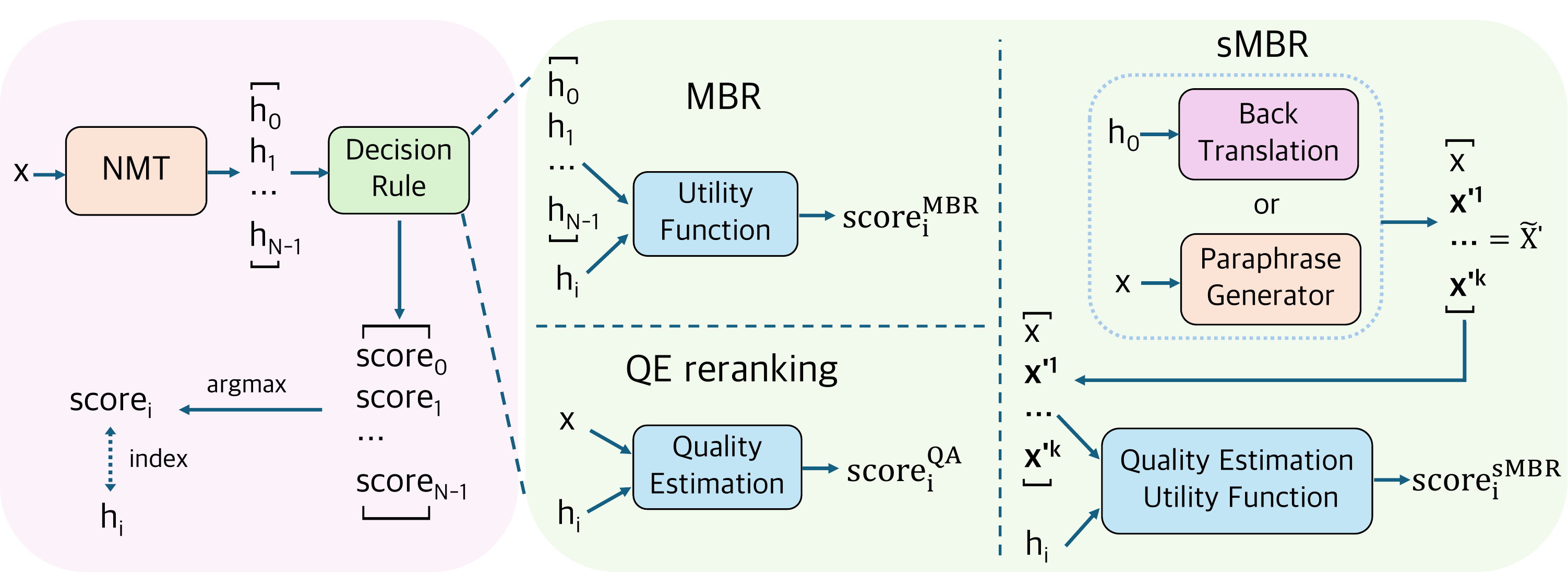}
    \caption{Overview of decoding methods in NMT. The diagram illustrates the process for MBR decoding, QE reranking, and the proposed sMBR decoding. It also shows two practices
    of sMBR: sMBR-BT and sMBR-PP. The figure demonstrates how the score used for selecting the final hypothesis is computed for each method.}
    \label{fig:overview}
\end{figure*}
\subsection{QE reranking}
Quality estimation (QE) is a task that aims to assess the quality of a translated sentence without reference translations but the original source sentence $x$. Recently, QE models have been employed to develop a new decoding method called QE reranking (\citealp{fernandes-etal-2022-quality}), which leverages QE models to rerank the candidate hypotheses. The main idea behind QE reranking is to select the hypothesis $h$ with the highest estimated quality with the QE model, rather than relying on the estimated probability $P_{mt}(h|x)$.

Formally, for a source ${x}$, a candidate hypothesis from the hypothesis space ${h \in \mathcal{C}}$, and QE function ${f_{\mathit{QE}}}$, QE reranking aims at finding a ${y^{\mathit{QE}}}$ that has the highest QE decision score ${score^{\mathit{QE}}_{h}}$:
\begin{align}
y^{\mathit{QE}} & = \mathop{\mathrm{argmax}}\limits_{h \in \mathcal{C}} {score^{\mathit{QE}}_{h}}
\\ & =  \mathop{\mathrm{argmax}}\limits_{h \in \mathcal{C}} {f_{\mathit{QE}}(x,h)}.
\end{align}
\section{Proposed Method}
In this section, we first introduce our proposed method, source-based MBR (sMBR) decoding. Then, we show that QE reranking is actually a special case of sMBR. Finally, we introduce two practices of sMBR: paraphrasing-based (sMBR-PP) and back-translation-based (sMBR-BT). See Figure~\ref{fig:overview} for a quick overview of how sMBR differs from standard MBR decoding and QE reranking.

\subsection{sMBR}\label{sec:sMBR}
In this subsection, we introduce MBR decoding using a novel method to calculate the utility, sMBR decoding. Then, we will show that QE reranking is actually a special case of sMBR.

We hypothesize that for a hypothesis, decision scores calculated by QE using only the hypothesis and sources are a good proxy for translation quality, since QE reranking can achieve promising performance using only a reference-free reranker (QE model) (\citealp{fernandes-etal-2022-quality}). We are therefore interested in calculating the utility of MBR decoding using only sources and a QE model, and call MBR decoding based on this idea \textbf{sMBR decoding}. In other words, in the sMBR, sources are used as ``support hypotheses'', and the QE model is used as a utility function.

Given that empirically better performance can be obtained by using more support hypotheses in the standard MBR decoding (\citealp{freitag-etal-2022-high}; \citealp{fernandes-etal-2022-quality}; \citealp{freitag-etal-2023-epsilon}), we would like to use more ``support hypotheses'' (sources) for sMBR as well. However, in the standard MBR decoding, there are usually multiple support hypotheses (i.e., $|\mathcal{S}|
\textgreater 1$), while we usually only have one source sentence. Therefore, we obtain additional ``support hypotheses'' for sMBR by considering other source language sentences that have the same meaning as the original source sentence.

Formally, let $P_{pp}(X'|X)$ be a paraphrasing distribution of source language sentences with the same meaning 
as the original source $x$, and $\widetilde{X'}$ be a finite sample from $P_{pp}(X'|X=x)$.
Then, for a reference-free QE utility function ${u(\cdot,\cdot})$,
sMBR looks for ${y^{\mathit{sMBR}}}$ that has the highest sMBR decision score ${score^{\mathit{sMBR}}_{h}}$
in the set of candidate hypotheses ${\mathcal{C}}$:
\begin{align}
y^{\mathit{sMBR}} 
& =  \mathop{\mathrm{argmax}}\limits_{h \in {\mathcal{C}}} {\mathbb{E}_{x'\in \widetilde{X'}}\left[u(x', h) \mid x \right]}
\\ & = \mathop{\mathrm{argmax}}\limits_{h \in \mathcal{C}} \sum_{x' \in \widetilde{X'}} u(x', h)P_{pp}(x'|x).
\label{def:sMBR_x}
\end{align}

Similarly to eq. \ref{def:MBR_appro}, 
it is also possible to approximate by assuming that all sources $x'$ have the same probability, which simplifies the calculation.
Formally, let ${K = |\widetilde{X'}|}$, 
then the objective of sMBR can be approximated as:
\begin{align}
y^{\mathit{sMBR}} & \approx \mathop{\mathrm{argmax}}\limits_{h \in \mathcal{C}} 
score^{\mathit{sMBR}}_{h},\\
score^{\mathit{sMBR}}_{h} & =
\frac{1}{K} \sum_{x' \in \widetilde{X'}}u(x', h).
\label{def:sMBR_appro}
\end{align}

Here, QE reranking is a special case of sMBR when ${K = 1}$, i.e., when only the original source is used. 
Unlike QE reranking, sMBR considers multiple quasi-sources (${K \textgreater 1}$), which are intended to serve a more diverse and representative utility.

\subsection{sMBR-PP and sMBR-BT}
We refer to the exact formulation presented in section~\ref{sec:sMBR} as sMBR-PP.
%
%
In addition, we 
study an alternative approach, which indirectly generates 
quasi-sources
by \textbf{b}ack-\textbf{t}ranslation, \textbf{sMBR-BT}.
Specifically, for an original source ${x}$, we first generat a translation ${h_0}$ using the forward translation model $P_{mt}(h|x)$ and then use ${h_0}$ as the input to a back-translation model to generate ${K}$ quasi-sources ${\{x'^1,x'^2,\ldots,x'^k\}}$. 
We then use the set ${\widetilde{X'} = \{x,x'^1,x'^2,\ldots,x'^k\}}$ of size $K+1$ for sMBR decoding. Note that in sMBR-BT, the input to the back-translation model is a single hypothesis, where we simply use the 
one with the
highest estimated probability. 
Then, we obtain ${K}$ quasi-sources by beam search with beam size ${K}$.

\vspace{\baselineskip}
\section{Experiments}
\subsection{Setup}
The details of NMT systems, decoding methods, and evaluation metrics are described.

\subsubsection{Data and models for NMT}
We consider two setups: the classic (encoder-decoder based transformer model trained from scratch on parallel corpora) and the LLM setup.

Our experiments were performed on the English to German (En$\rightarrow$De), English to Russian (En$\rightarrow$Ru), and Chinese to English (Zh$\rightarrow$En), as COMET and COMET-QE on them proved to be highly correlated with human judgments at the segment level (\citealp{rei-etal-2022-comet}). We use \code{generaltest2023} (\citealp{kocmi-etal-2023-findings}) as the test set for each translation direction.
\paragraph{Classic setup}
The classic setup does not include the Zh$\rightarrow$En due to computing resource limitations.  
For both En$\rightarrow$De and En$\rightarrow$Ru in the classic setup, we use \code{newstest2017-2019} as the development set.

We further consider both high-resource and low-resource sub-setups. 
For the high-resource setup, we use Facebook FAIR's WMT19 news translation task submission (\citealp{ng-etal-2019-facebook}). 
For low resource setup, training data consists of 0.44M and 0.38M parallel sentences for En$\rightarrow$De and En$\rightarrow$Ru, respectively. These systems allow us to assess the performance in a data-constrained scenario. See 
Appendix~\ref{sec:classic_setup} for more details.
\paragraph{LLM setup}
We use \code{TowerInstruct-13B} (\citealp{Alves2024TowerAO}), a state-of-the-art LLM for translation-related tasks as NMT model. We prompt LLM to perform zero-shot translation. See 
Appendix~\ref{sec:tower} for more details.

\subsubsection{Decoding methods}
We employ four approaches for hypothesis generation: beam search, ancestral sampling, top-k sampling, and epsilon sampling. Beam search has been widely used for MBR decoding in the past (\citealp{kumar2004minimum}; \citealp{stahlberg-etal-2017-neural}; \citealp{DBLP:journals/corr/ShuN17}; \citealp{blain-etal-2017-exploring}), while ancestral sampling has gained popularity in recent work (\citealp{eikema-aziz-2020-map}; \citealp{freitag-etal-2022-high}; \citealp{eikema-aziz-2022-sampling}). We include top-k sampling and epsilon sampling since we find that they yield better performance than ancestral sampling.

Both for the classic and LLM setups
, we evaluate the following decision rules:
\begin{itemize}
    \item MAP: A widely used rule that selects the hypothesis with the highest estimated probability.

    \item MBR: MBR decoding based on COMET, using the \code{Unbabel/wmt22-comet-da}\footnote{\url{huggingface.co/Unbabel/wmt22-comet-da}} model. It calculates decision scores for candidate hypotheses using the approach described in \ref{sec:mbr}, and then select the highest score one.

    \item QE reranking: A special form of MBR, which calculates decision scores with the quality estimation model ${f_{\mathit{QE}}(\cdot,\cdot)}$. Since COMET does not support reference-free quality estimation, we use COMET-QE (\code{Unbabel/wmt22-cometkiwi-da}\footnote{\url{huggingface.co/Unbabel/wmt22-cometkiwi-da}}) as the utility function.

    \item sMBR: Source-based MBR decoding. We evaluate its two practices: sMBR-PP and sMBR-BT. The same QE model used for QE reranking is employed as the utility function. 
    For sMBR-PP, 
    we fine-tuned a unique \code{T5-large} (\citealp{chung2024scaling}) model as a paraphrase generator for each source language, which was fine-tuned on paraphrase data created through back-translation. We detail the sMBR-PP implementation in Appendix~\ref{sec:smbr_details}.
    As the back-translation model of sMBR-BT, in the classic setup, we again use Facebook FAIR's WMT19 news translation task submission in the high resource setup; in the low resource setup, we use the same model architecture and data as in the forward-translation model.
    For the LLM setup, we reused the LLM (\code{TowerInstruct-13B}) as a back-translation model.
\end{itemize}

Our baseline is a MAP based on beam search with a beam size of 5 both for the classic and LLM setups, since we found that larger beams do not lead to better performance. 
Except for the baseline, we use \textbf{400} candidate hypotheses for the classic setup unless otherwise specified, as we find that more hypotheses have limited gains in performance but result in higher costs.\footnote{Appendix~\ref{sec:candidate} shows the impact of candidate hypotheses number on the metrics.} For the LLM setup, we only use \textbf{128} hypotheses because generating more hypotheses leads to too much computing.

For MBR decoding, we tried two setups: (1) using the same set of hypotheses as candidate hypotheses and support hypotheses; and (2) using QE reranking to filter the set of support hypotheses to a smaller size that matches the size of the set of support hypotheses for sMBR. For sMBR-PP and sMBR-BT, we study case using 16 quasi-sources, as adding more did not yield further gains.

More details are provided in Appendix~\ref{sec:setup}.

\subsubsection{Evaluation metrics}
We use automatic evaluation metrics, including BLEU (\citealp{post-2018-call}), XCOMET (\code{XCOMET-XXL}) (\citealp{Guerreiro2023xCOMETTM}), and MetricX (\code{MetricX-24-Hybrid-XXL}) (\citealp{metricx24}) to evaluate different methods. 
Our choice of XCOMET and MetricX is motivated by two key factors: 1) they are state-of-the-art neural metrics (\citealp{wmt24_metrics}), which correlate with human judgments even when evaluating NMT systems that use neural metric-based reranking (\citealp{kovacs-etal-2024-mitigating}); 2) they were trained on Multidimensional Quality Metrics (MQM) data (\citealp{Guerreiro2023xCOMETTM}; \citealp{metricx24}), while the COMET series, 
on which MBR and sMBR were directly optimized,
were trained only on Direct Assessments (DA) data (\citealp{rei-etal-2022-comet}).
Given the limited correlation between MQM and DA (\citealp{freitag-etal-2021-experts}), we expect XCOMET and MetricX to provide a more independent assessment, as they are less likely to be biased towards the COMET-optimized MBR and sMBR methods.

For all evaluations, we use reference translations unless otherwise mentioned.
We perform significance tests using paired bootstrap resampling (\citealp{koehn-2004-statistical}).

\subsection{Results}\label{sec:results}

\begin{table*}[!t]
\scalebox{0.92}{
\centering
\begin{tabular}{lrrllllllll}
\toprule
\textbf{Decoding method } & & & \multicolumn{3}{c}{\textbf{En$\rightarrow$De}} & \multicolumn{3}{c}{\textbf{En$\rightarrow$Ru}} \\
\cmidrule(r){4-6} \cmidrule(l){7-10}
w/ beam search & $|\mathcal{C}|$ & $|\mathcal{S}|$ & \textbf{BLEU$\uparrow$} & \textbf{XCOMET$\uparrow$} & \textbf{MetricX$\downarrow$} & \textbf{BLEU$\uparrow$} & \textbf{XCOMET$\uparrow$} & \textbf{MetricX$\downarrow$} \\
\midrule
\multicolumn{8}{r}{\textbf{High Resource (55.4M and 52.0M training data)}} \\
\midrule
 MAP& 5 & -- &34.80 & 84.89 &3.63&27.82&82.90&5.36\\
 MBR& 5 & 5 &34.97&85.12&3.58&27.35&83.62&5.01\\
\hdashline
MAP & 400 & -- & 34.30 & 85.10 & 3.69&\textbf{23.05}&82.99&6.49 \\
QE reranking & 400 & 1 & \textbf{35.23} & 86.48 & 3.22&22.80&86.20&4.27 \\
MBR & 400 & 400 & 34.83 & 85.74 & 3.50&22.81&84.95&5.17\\
 MBR& 400 & 17 & 34.93 & 85.88 & 3.34&\textbf{23.05}&85.00&5.37\\
sMBR-PP & 400 & 17 &34.81&\textbf{86.73}\textsuperscript{†} &\textbf{3.09}\textsuperscript{††} &22.91&\textbf{86.52}\textsuperscript{††}&\textbf{4.14}\textsuperscript{†}\\
sMBR-BT & 400 & 17 & 33.80 & 86.17 & 3.33&22.36&84.99&4.65 \\
\midrule
\multicolumn{8}{r}{\textbf{Low Resource (0.44M and 0.38M training data)}} \\
\midrule
 MAP &5&--& 11.44&60.29&12.43 &17.44&65.85&10.98\\
 MBR& 5 & 5 &12.56&60.33&12.16 &17.35&66.95&10.51\\
\hdashline
MAP &400&--& 9.78&62.58&12.24 &17.51&66.36&10.85\\
QE reranking &400&1& \textbf{13.63}&65.63&10.34&17.71&74.66&7.81 \\
MBR &400&400& 12.66 &63.79&11.05&17.56&70.69&9.01\\
 MBR& 400 & 17 &11.75&64.53&11.11&\textbf{18.17}\textsuperscript{††}&71.10&9.07 \\
sMBR-PP & 400& 17& 13.49&\textbf{66.36}\textsuperscript{††}&\textbf{10.19}\textsuperscript{†}&17.95&\textbf{74.96}\textsuperscript{††}&\textbf{7.76} \textsuperscript{†} \\
sMBR-BT &400&17& 9.66&63.77&11.68&16.87&69.69& 9.35\\
\bottomrule
\end{tabular}}
\caption{Compares decision rules in the classic setup. $|\mathcal{C}|$ and $|\mathcal{S}|$ indicate the number of candidate hypotheses and supportive hypotheses, respectively. For sMBR, we used $|\mathcal{S}|=$ 17 support hypotheses (1 original source + 16 quasi-sources). We performed paired bootstrap resampling over all methods against QE reranking; † and †† indicate significantly better than QE reranking within groups ($p < 0.05$ and $p < 0.01$, respectively; Multiple testing correction is not applied). The best in each group is marked in bold.}
\label{tab:overall}
\end{table*}

\begin{table*}[h]
\scalebox{0.92}{
\centering
\begin{tabular}{lrrllllllll}
\toprule
\textbf{Decoding method } & & & \multicolumn{3}{c}{\textbf{En$\rightarrow$De}}  & \multicolumn{3}{c}{\textbf{Zh$\rightarrow$En}} \\
\cmidrule(r){4-6} \cmidrule(l){7-10}
~ & $|\mathcal{C}|$ & $|\mathcal{S}|$ & \textbf{BLEU$\uparrow$} & \textbf{XCOMET$\uparrow$} & \textbf{MetricX$\downarrow$} & \textbf{BLEU$\uparrow$} & \textbf{XCOMET$\uparrow$} & \textbf{MetricX$\downarrow$} \\
\midrule
w/ epsilon sampling & \multicolumn{8}{c}{\textbf{\textit{TowerInstruct-13B}}} \\
\midrule
 MAP& 5 & -- & 30.24&86.06 & 3.32 &20.73 &88.15 &2.41 \\
 MBR& 5 & 5 & 28.10& 87.30 &2.95 &19.74 &88.96 & 2.20 \\
\hdashline
MAP & 128& -- & \textbf{32.64}\textsuperscript{††}&86.43 &3.22&23.12\textsuperscript{††} &89.14 &2.20 \\
QE reranking & 128 & 1 & 29.40& 88.76 &2.56 &19.88 &90.64 & 1.89 \\
MBR & 128 & 128 & 29.84&89.19\textsuperscript{†} & 2.46\textsuperscript{††} &22.01\textsuperscript{††} & 90.39& 1.90 \\
 MBR& 128 & 17 & 31.93\textsuperscript{††}& 88.83 & 2.60 &\textbf{23.34}\textsuperscript{††}& 90.43 & 1.87 \\
sMBR-PP & 128 & 17 & 27.19 & \textbf{89.47}\textsuperscript{††}&  \textbf{2.44}\textsuperscript{††} &19.87 &\textbf{90.70}\textsuperscript{†} &  \textbf{1.87}\\
sMBR-BT &128 & 17 & 28.73& 89.04 &2.50 &19.77 & 90.38& 1.98 \\
\bottomrule
\end{tabular}}
\caption{Compares decision rules in the {LLM setup}. The meaning of table elements is the same as Table \ref{tab:overall}.}
\label{tab:smbr_tower_eps}
\end{table*}

We analyze the performance by observing the results of the automatic evaluation metrics.
Due to space constraints, in this subsection, we show in the main text only the results of beam search based results in the classic setup and epsilon sampling based results in the LLM setup; other results are included in Appendix~\ref{sec:AS}.
\paragraph{Classic setup}Table \ref{tab:overall} highlights the effectiveness of sMBR decoding in the classic setup with beam search.
Regarding XCOMET and MetricX, sMBR-PP significantly outperforms QE reranking, proving the validity of our extension to QE reranking.
The results of the experiments based on sampling methods are shown in
Appendix~\ref{sec:AS}, where similar gains to those based on beam search can be observed.
The sMBR-PP outperforms the standard MBR on the neural metric, regardless of whether the standard MBR can use the full 400 support hypotheses or only 17.
Thus, we conclude that the sMBR-PP
outperforms QE reranking and 
the standard MBR in the classic setup.
\paragraph{LLM setup}Table~\ref{tab:smbr_tower_eps} shows the results of epsilon sampling in the LLM setup.
We observe that sMBR-PP can still significantly outperform QE reranking regarding XCOMET and MetricX. 
The XCOMET and MetricX of sMBR-PP are comparable to the standard MBR and sometimes outperform standard MBR.
As with the results of the classic setup, a gain similar to that based on epsilon sampling can be observed in Appendix \ref{sec:AS}.
Thus, we conclude that sMBR-PP still significantly outperforms QE reranking and is competitive with the standard MBR in the LLM setup.

The performance of sMBR-PP relative to standard MBR differs between the two setups. The standard MBR shows performance similar to that of the sMBR-PP in the LLM setup. We believe this is due to the better quality of the support hypothesis generated by LLM, which leads to a higher approximation accuracy in eq. \ref{def:MBR_appro}.

Compared to QE reranking, sMBR-BT shows gains regarding XCOMET and MetricX in the LLM setup, but even lower metrics than QE reranking in the classic setup.
We investigate this reason in Appendix \ref{sec:anal_syn}. 

\section{Discussion}
In this section, we discuss the effectiveness of sMBR-PP and the mechanism behind it through some experiments.
Due to space constraints, we discuss the efficiency of sMBR-PP in Appendix \ref{sec:speed_with_mbr}, where we show that sMBR-PP is much slower than QE reranking and MBR with simple optimization.
\subsection{Effects of increasing sources}
\begin{table}[t]
\scalebox{0.88}{
\centering
\begin{tabular}{cccccc}
\toprule
$|\mathcal{S}|$ & 1 & 6 & 11 & 17 & 33\\
\midrule
XCOMET $\uparrow$ &86.48& 86.65&86.73&86.73&86.74\\
MetricX $\downarrow$ &3.22&3.12&3.10& 3.09&3.09\\
\bottomrule
\end{tabular}}
\caption{Impact of increasing sources for sMBR-PP: The number of sources is positively correlated with the evaluation metrics. $|\mathcal{S}|=1+K$, i.e., (\# of the original source) + (\# of  quasi-sources). 
Candidate hypotheses were generated by beam search.
When
$|\mathcal{S}|=1$,
sMBR-PP is QE reranking.}
\label{tab:scab}
\end{table}
Since sMBR is an extension to QE reranking by increasing the number of sources, we first investigate the impact of increasing the number of sources on the performance of sMBR-PP. We focus on the En$\rightarrow$De high resource setup of the classic setup and evaluate with neural metrics. Table~\ref{tab:scab} presents the results, demonstrating a positive correlation between source number and evaluation metrics. This observation again shows that our extensions to QE reranking are effective. In addition, the increase in synthesis sources from 16 to 32 does not result in further gains, which we hypothesize is due to the inability of the paraphrase generator to achieve the generation of up to 32 generative high-quality synthesis sources.
\subsection{
Qualities of
quasi-sources}\label{sec:anal_syn}
\begin{table}[t]
\scalebox{0.95}{
\centering
\begin{tabular}{lccc}
\toprule
~ & Self-BLEU$\downarrow$ & Semantic Similarity$\uparrow$ \\
\midrule
sMBR-PP & 41.68 &  94.32\\
sMBR-BT & 48.25 &  94.53\\
\bottomrule
\end{tabular}}
\caption{Analyzing of quasi-sources: analyzed on the En$\rightarrow$De \code{generaltest2023}, high resource. Lower Self-BLEU means richer surface diversity; higher semantic similarity means closer semantics to the original source.}
\label{tab:synt_src}
\end{table}
To understand the properties of the quasi-sources of sMBR, we analyzed them in terms of surface diversity and semantic similarity with the original source. Surface diversity was measured using Self-BLEU (\citealp{10.1145/3209978.3210080}), while semantic similarity was assessed by cosine distance between sentence embeddings\footnote{We use \url{huggingface.co/sentence-transformers/all-mpnet-base-v2}}.

The results presented in Table~\ref{tab:synt_src} reveal that the quasi-sources generated by sMBR-PP exhibit much lower Self-BLEU scores compared to those produced by sMBR-BT, indicating greater surface diversity. On the other hand, the scores of the two in terms of semantic similarity are close, implying that both generated quasi-sources do not deviate too much from the original source's semantics. We hypothesize that the poor performance of sMBR-BT in the classic setup can be attributed to the limited surface diversity of its quasi-sources.

\label{sec:understand_smbr}
\begin{table}[t]
\centering
\begin{tabular}{lccccc}
\toprule
$|\mathcal{S}|$ & 1 & 6 & 11 & 17 \\
\midrule
Ave. QE
& 81.28 & 80.58 & 80.57 & 80.54 \\
\bottomrule
\end{tabular}
\caption{Average QE scores with the original source: QE scores are negatively correlated with source number. The analysis was performed in the En$\rightarrow$De high resource setup on sMBR-PP. Candidate hypotheses were generated by beam search.}
\label{tab:QE}
\end{table}
\subsection{Why sMBR-PP works}
The results shown in Section \ref{sec:results} demonstrate that sMBR-PP is significantly better than QE reranking in terms of neural metrics.
We discuss the mechanism behind producing these gains.
We hypothesize that the QE model used in QE reranking is overly sensitive to the specific phrasing and structure of the original source, leading to an over-reliance on a single source that could negatively impact performance. In contrast, aggregating QE scores across multiple sources in sMBR decoding is expected to provide more robust QE. We observed that as the number of sources increases in sMBR decoding, the average QE score between the selected translations and the original source sentence decreases (Table~\ref{tab:QE}). This suggests that sMBR decoding no longer relies solely on QE with respect to the original source. We conjecture that this is because sMBR decoding tends to select hypotheses that are more generally applicable to different source variants.
\section{Related Work}
MBR decoding has been used in speech recognition (\citealp{Stolcke1997ExplicitWE}; \citealp{goel2000minimum}), word alignment (\citealp{kumar-byrne-2002-minimum}), and statistical machine translation (\citealp{kumar2004minimum}; \citealp{tromble-etal-2008-lattice}). Recently, some works have re-explored the application of MBR decoding in NMT and demonstrated promising results (\citealp{stahlberg-etal-2017-neural}; \citealp{DBLP:journals/corr/ShuN17}; \citealp{eikema-aziz-2020-map}; \citealp{eikema-aziz-2022-sampling}). These works have shown that MBR decoding can help overcome some of the limitations of MAP.

In past work, MBR decoding is usually based on beam search to generate candidate hypotheses (\citealp{stahlberg-etal-2017-neural}; \citealp{DBLP:journals/corr/ShuN17}). Recently, \citet{eikema-aziz-2020-map} proposed sampling-based MBR decoding and found that the samples from the model are faithful to the training data statistics, while the beam search is not. \citet{freitag-etal-2022-high} further explored the impact of the generation method on the performance.

In terms of utility functions, past work has primarily used surface-based metrics such as BLEU and BEER (\citealp{stanojevic-simaan-2014-fitting}). However, these metrics have limited correlation with human judgments (\citealp{mathur-etal-2020-tangled}; \citealp{freitag-etal-2023-results}). Recently, a trend has been to combine advanced neural metrics with MBR decoding, such as COMET and BLEURT. These works demonstrate that neural metrics-based MBR can improve performance in human evaluations (\citealp{freitag-etal-2022-high}; \citealp{fernandes-etal-2022-quality}; \citealp{freitag-etal-2023-epsilon}). However, they are also limited by the high cost, as MBR decoding has a secondary cost for the number of candidate hypotheses. \citet{eikema-aziz-2022-sampling} investigated decoupling candidate and support hypotheses, enabling the exploration of more potential candidate hypotheses within a limited computational budget.
In addition, some recent work has focused on improving the efficiency of the MBR decoding (\citealp{cheng-vlachos-2023-faster}; \citealp{vamvas-sennrich-2024-linear}; \citealp{deguchi-etal-2024-centroid}).

On the other hand, some work has found that models used to assess the quality of NMT systems (i.e., quality estimation) can perform well even in the absence of a reference (\citealp{rei-etal-2021-references}; \citealp{rei-etal-2022-cometkiwi}; \citealp{rei-etal-2023-scaling}). \citet{fernandes-etal-2022-quality} explored the direct use of quality estimation models as rerankers for NMT and showed promising results, referred to as QE reranking.

\section{Conclusions and Future Work}
In this work, inspired by QE reranking, we propose sMBR decoding, which uses sources and QE model to calculate the utility, the first practical method to solely rely on sources as ``support hypotheses'' in MBR decoding.
Experimental results (Section~\ref{sec:results}) show that sMBR decoding outperforms QE reranking and the standard MBR decoding.

Despite its limitations, such as the challenge of generating quasi-sources, sMBR represents a significant step forward in MBR decoding. By breaking with the tradition of approximating true utility using only the average of utilities with respect to other hypotheses, sMBR opens up new possibilities for future research.

Our analysis in Appendix~\ref{sec:smbr_gpt} indicates that using a more powerful paraphrase generator, such as GPT4 (\citealp{Achiam2023GPT4TR}), for sMBR-PP shows promise for further performance improvements. The analysis in Appendix~\ref{sec:dbs} suggests that using Diverse Beam Search (\citealp{dbs}) for sMBR-BT has the potential to enhance performance. Therefore, we plan to explore these methods for generating quasi-sources in our future work.
In addition to methods for generating quasi-sources, in our future work we will continue to investigate broadening the boundaries of ``support hypotheses'' to include sentences in languages other than the source and target.
\section{Limitations}
\label{sec:limitations}
While our proposed sMBR decoding approach shows promising results, it has some limitations.

Firstly, reranking methods that directly optimize evaluation metrics may ``overfit'' to those metrics, causing the optimized metrics 
to become unreliable (\citealp{kovacs-etal-2024-mitigating}). We mitigate this problem by using automatic metrics that are likely to have low correlation with the metrics that are directly optimized. However, since they still have common parts in the training data, this measure may not completely avoid the problem of unreliable metrics. Moreover, extending QE reranking with a quasi-source may exacerbate this problem, as it may result in the final selection being favored by the QE metric itself but translations that are unrelated to the source (referred to as ``universal translations'' by \citet{yan-etal-2023-bleurt}). Therefore, our conclusion that sMBR-PP outperforms QE reranking and standard MBR decoding may be questioned. Human evaluation can mitigate this issue but is costly and time-consuming.

Secondly, generating high-quality quasi-source sentences remains a challenge. We explored two methods based on paraphrasing and back-translation, but the back-translation approach did not consistently improve reranking performance. This suggests that further research is needed to identify more effective techniques for generating diverse, representative quasi-sources.

Finally, we have only tested the proposed method in a limited number of translation directions and domains. However, not all language pairs have well-performing quality estimation models available.
In the case of some language pairs, this may lead to a questioning of one of our basic hypothesis, i.e., the quality estimation model is a good proxy for the true utility. Therefore, the effectiveness of sMBR in a wider range of settings remains an open question.

\section*{Acknowledgements}
We are grateful to the reviewers in the past rounds of the ACL Rolling Review, whose comments have allowed us to continually improve our work.

\bibliography{custom}
\newpage
\appendix

\section{Additional details of sMBR-PP}
\label{sec:smbr_details}
For sMBR-PP, we generally use the same paraphrase generator for both the classic and LLM setups for the same source language.

When the source is English, for the paraphrase generator used in sMBR-PP, its specific model is \code{google/flan-t5-large}\footnote{\url{huggingface.co/google/flan-t5-large}}.
This model is trained for instruction following and thus works out-of-the-box for paraphrase generation. However, we found its performance to be rather poor, thus we chose to fine-tune it.

The fine-tuning training data consists of a publicly available paraphrase generation dataset, PAWS (\citealp{paws2019naacl}), concatenated with a dataset we created. The dataset we created is based on En-De's News-Commentary parallel corpus\footnote{\url{data.statmt.org/news-commentary/v18.1/}} and uses machine translation to create paraphrased sentences. Specifically, we first input German sentences from the parallel corpus into the De$\rightarrow$En NMT model, and then paired its output with English sentences from the original parallel corpus to compose the samples in the paraphrase generation dataset. We use the De$\rightarrow$En model from Facebook FAIR’s WMT19 news
translation task submission (\citealp{ng-etal-2019-facebook}). We use a semantic similarity-based approach to estimate the quality of the dataset we created, and then filter out sentence pairs with low similarity. We use the \code{sentence-transformers/all-mpnet-base-v2}\footnote{\url{huggingface.co/sentence-transformers/all-mpnet-base-v2}} model to compute the similarity between paraphrase pairs and filter out sentence pairs with a similarity of 0.88 or less. In the end, the training data for the model consisted of a total of about 339.2K paraphrased sentence pairs, of which about 317.4K came from the data we created and about 21.8K came from PAWS.

For the training of this model, we used the AdamW optimizer (\citealp{DBLP:conf/iclr/LoshchilovH19}) with a learning rate of 3e-4, weight decay of 0.0, and a batch size of 1536 examples, trained with fp32 full precision (This is because we found that the \code{flan-t5} series is prone to training failure at fp16 precision). 
We set the maximum number of training epochs to 10. 
We randomly separate 3K sentence pairs from the dataset as the development set, and then select the checkpoints with the lowest loss on the development set. 

When the source is Chinese, we follow a similar procedure as above. The difference is that the base model is \code{mT5-large} (\citealp{Xue2020mT5AM}) and the training data only includes the dataset created by using \code{TowerInstruct-13B} to perform reverse translation on the News-Commentary parallel corpus. We use \code{lier007/xiaobu-embedding-v2}\footnote{\url{https://huggingface.co/lier007/xiaobu-embedding-v2}} to calculate the cosine similarity between the rephrase and the original sentence, and filter out samples with a similarity below 0.925.    

In the inference phase of generating paraphrases, we used epsilon sampling (\citealp{hewitt-etal-2022-truncation}) (epsilon = 0.02), as we found that this setup balances the diversity and quality of the synthesized sources well.
Training and inference were done on a single NVIDIA H100.

We used the following input and output format during training and inference: 

Source (input to the encoder):
\begin{center}
\code{\{source\_text\}}
\end{center}

Target (input to the decoder):
\begin{center}
\code{\{target\_text\}}
\end{center}

\section{Additional details on decoding and training of low-resource NMT models}
\label{sec:setup}
We completed training of the NMT low-resource model, and all decoding experiments on a machine with 4 NVIDIA RTX A6000. In the hypothesis generation phase, we used CTranslate2\footnote{\url{https://github.com/OpenNMT/CTranslate2}} to generate hypotheses because of its efficiency.

For the training of low-resource NMT models, we use the fairseq (\citealp{ott2019fairseq}) tool. We use base size transformer (\citealp{Vaswani2017AttentionIA}) architectures with a dropout rate of 0.3. And train for a maximum of 100 epochs at full fp16 precision. we select the checkpoints with the highest BLEU on the development set. We use adam (\citealp{DBLP:journals/corr/KingmaB14}) optimizer with an initial learning rate of 1e-3, weight decay of 1e-4, a warm-up step of 4000, and batch size is 1e5 tokens. We build vocabulary of size 32000 with Byte-Pair Encoding (\citealp{sennrich-etal-2016-neural}) using the sentencepiece (\citealp{kudo-richardson-2018-sentencepiece}) tool. The vocabulary is shared between source and target languages.
\section{Additional experimental results}
\label{sec:AS}
In this section, we present additional experimental results that could not be included in the main text due to space constraints. For the classic setup, this section includes the results of the generation methods other than beam search; for the LLM setup, this section includes not only the results of the generation methods other than Epsilon sampling, but also the complete experimental results of En->Ru.

In addition to beam search and epsilon sampling, we attempted to use top-k sampling and ancestral sampling to generate hypotheses. Unlike top-k sampling, ancestral sampling the entire vocabulary for each time step in autoregressive decoding without any pruning. The results of the experiments are shown in Table~\ref{tab:topk}, \ref{tab:smbr_eps}, \ref{tab:smbr_as}, \ref{tab:smbr_tower}, and \ref{tab:smbr_tower_ru}. 
Compared to other generation methods, ancestral sampling performs poorly on both surface-based and neural metrics. Among the sampling-based methods, epsilon sampling performs best, which is consistent with the findings of \citealp{freitag-etal-2023-epsilon}.

We used ${k = 10}$ for top-k sampling and epsilon ${= 0.02}$ for epsilon sampling (\citealp{freitag-etal-2023-epsilon}). Due to implementation issues with some CUDA programming, we do not consider epsilon sampling with low resource setup.  

In conclusion, similar to the experimental results based on beam search and epsilon sampling, the significantly boosted neural metrics demonstrate that sMBR-PP significantly outperforms QE reranking.
However, improvements in neural metrics do not always lead to gains in surface-based metrics and even lead to deterioration compared to the baseline, especially when using sampling-based hypothesis generation. One possible explanation is that sampling leads to more diverse hypotheses, making it easier to generate candidates hypotheses that would lead to higher neural metrics but not favored by BLEU. Unfortunately, sMBR decoding does not consistently mitigate this issue compared to QE reranking, suggesting potential limitations in the utility functions.
\begin{table*}[!t]
\scalebox{0.92}{
\centering
\begin{tabular}{lrrllllllll}
\toprule
\textbf{Decoding method } & & & \multicolumn{3}{c}{\textbf{En$\rightarrow$De}} & \multicolumn{3}{c}{\textbf{En$\rightarrow$Ru}} \\
\cmidrule(r){4-6} \cmidrule(l){7-10}
w/ top-k sampling & $|\mathcal{C}|$ & $|\mathcal{S}|$ & \textbf{BLEU$\uparrow$} & \textbf{XCOMET$\uparrow$} & \textbf{MetricX$\downarrow$} & \textbf{BLEU$\uparrow$} & \textbf{XCOMET$\uparrow$} & \textbf{MetricX$\downarrow$} \\
\midrule
\multicolumn{8}{r}{\textbf{High Resource (55.4M and 52.0M training data)}} \\
\midrule
 MAP& 5 & -- & 24.36&70.80&7.81&17.98&66.34&8.69\\
 MBR& 5 & 5 & 23.19&71.56&7.46&16.26&68.18&7.78\\
\hdashline
MAP & 400 & -- &24.42&69.55&8.11&21.31\textsuperscript{††}&77.83&7.99 \\
QE reranking & 400 & 1 &26.69&81.99&4.46&17.94&84.34&4.38 \\
MBR & 400 & 400 & 25.13&78.51&5.57&19.26\textsuperscript{††}&80.86&5.18\\
 MBR& 400 & 17 &\textbf{27.99}\textsuperscript{††}&80.28&4.89&\textbf{22.21}\textsuperscript{††}&81.55&5.70\\
sMBR-PP & 400 & 17 &25.74&\textbf{82.14}&\textbf{4.37}\textsuperscript{†}&17.30&\textbf{84.77}\textsuperscript{††}&\textbf{4.20}\textsuperscript{††}\\
sMBR-BT & 400 & 17 &25.01&77.29&6.00&16.75&79.54&5.80\\
\midrule
\multicolumn{8}{r}{\textbf{Low Resource (0.44M and 0.38M training data)}} \\
\midrule
 MAP& 5 & -- & 7.91&53.11&14.39&14.36&61.89&12.32\\
 MBR& 5 & 5 & 8.48&53.62&14.09&13.76&62.77&11.68\\
\hdashline
MAP & 400 & -- &6.05&58.64&13.48&16.93\textsuperscript{††}&65.55&11.12\\
QE reranking & 400 & 1 & \textbf{10.76}&62.47&11.01&15.06&73.50&7.94 \\
MBR & 400 & 400 & 10.02&60.80&11.99&15.48&69.61& 9.10\\
 MBR& 400 & 17 &10.13&61.44&11.74&\textbf{17.36}\textsuperscript{††}&69.95&9.32\\
sMBR-PP & 400 & 17 &10.36& \textbf{63.35}\textsuperscript{††}&\textbf{10.90}\textsuperscript{††}&15.11&\textbf{73.71}\textsuperscript{†}&\textbf{7.88}\textsuperscript{†}\\
sMBR-BT & 400 & 17 &5.64&59.96&12.86&14.70&68.22&9.78\\
\bottomrule
\end{tabular}}
\caption{Compares sMBR with other decision rules for En$\rightarrow$De and En$\rightarrow$Ru in the \textbf{classic setup}. $|\mathcal{C}|$ and $|\mathcal{S}|$ indicate the number of candidate hypotheses and supportive hypotheses, respectively. For sMBR, we used $|\mathcal{S}|=$ 17 ``support hypotheses'' (1 original source + 16 quasi-sources). We performed paired bootstrap resampling over all methods against QE reranking; † and †† indicate significantly better than QE reranking within groups ($p < 0.05$ and $p < 0.01$, respectively; Multiple testing correction is not applied). The best in each group is marked in bold.}
\label{tab:topk}
\end{table*}

\begin{table*}[!t]
\scalebox{0.92}{
\centering
\begin{tabular}{lrrllllllll}
\toprule
\textbf{Decoding method } & & & \multicolumn{3}{c}{\textbf{En$\rightarrow$De}} & \multicolumn{3}{c}{\textbf{En$\rightarrow$Ru}} \\
\cmidrule(r){4-6} \cmidrule(l){7-10}
w/ ancestral sampling & $|\mathcal{C}|$ & $|\mathcal{S}|$ & \textbf{BLEU$\uparrow$} & \textbf{XCOMET$\uparrow$} & \textbf{MetricX$\downarrow$} & \textbf{BLEU$\uparrow$} & \textbf{XCOMET$\uparrow$} & \textbf{MetricX$\downarrow$} \\
\midrule
\multicolumn{8}{r}{\textbf{High Resource (55.4M and 52.0M training data)}} \\
\midrule
 MAP& 5 & -- & 10.00&35.78&18.68&14.05&61.37&12.22\\
 MBR& 5 & 5 &6.80& 30.50&20.09&12.40&58.33&12.60\\
\hdashline
MAP & 400 & -- &11.60&51.08&14.71 &\textbf{21.83}\textsuperscript{††}&79.61&7.06\\
QE reranking & 400 & 1 & 16.98 &61.07&11.98&19.12&82.12&5.12\\
MBR & 400 & 400 &10.48&45.41 &15.88&17.64&78.47&6.05\\
 MBR& 400 & 17 &\textbf{18.23}\textsuperscript{††}&58.37&12.49&21.44&81.47&5.29\\
sMBR-PP & 400 & 17 &17.09&\textbf{61.29}& \textbf{11.97}&18.55&\textbf{82.22}&\textbf{5.11}\\
sMBR-BT & 400 & 17 &14.73&56.94&12.95&19.40&80.42&5.74\\
\midrule
\multicolumn{8}{r}{\textbf{Low Resource (0.44M and 0.38M training data)}} \\
\midrule
 MAP& 5 & -- & 5.29&39.33&18.35&11.13&49.98&16.13\\
 MBR& 5 & 5 & 5.85&36.64&18.92&8.47&46.38&16.73\\
\hdashline
MAP & 400 & -- &3.72&51.78&15.35&15.23\textsuperscript{††}&62.82&12.00 \\
QE reranking & 400 & 1 & 6.67&54.18&14.14&14.22&68.13&9.96 \\
MBR & 400 & 400 & \textbf{7.67}\textsuperscript{††}&47.81&15.70&12.47&62.45&11.44\\
 MBR& 400 & 17 &6.99&53.16&14.49&15.13\textsuperscript{††}&66.23&10.47\\
sMBR-PP & 400 & 17 &6.67&\textbf{54.85}\textsuperscript{††}&\textbf{14.11}&14.08&\textbf{68.41}\textsuperscript{†}&\textbf{9.88}\textsuperscript{†}\\
sMBR-BT & 400 & 17 &3.62&51.60&15.19&13.90&64.30&11.33\\
\bottomrule
\end{tabular}}
\caption{Compares sMBR with other decision rules for En$\rightarrow$De and En$\rightarrow$Ru in the \textbf{classic setup}. $|\mathcal{C}|$ and $|\mathcal{S}|$ indicate the number of candidate hypotheses and supportive hypotheses, respectively. For sMBR, we used $|\mathcal{S}|=$ 17 ``support hypotheses'' (1 original source + 16 quasi-sources). We performed paired bootstrap resampling over all methods against QE reranking; † and †† indicate significantly better than QE reranking within groups ($p < 0.05$ and $p < 0.01$, respectively; Multiple testing correction is not applied). The best in each group is marked in bold.}
\label{tab:smbr_as}
\end{table*}

\begin{table*}[!t]
\scalebox{0.92}{
\centering
\begin{tabular}{lrrllllllll}
\toprule
\textbf{Decoding method } & & & \multicolumn{3}{c}{\textbf{En$\rightarrow$De}} & \multicolumn{3}{c}{\textbf{En$\rightarrow$Ru}} \\
\cmidrule(r){4-6} \cmidrule(l){7-10}
w/ epsilon sampling & $|\mathcal{C}|$ & $|\mathcal{S}|$ & \textbf{BLEU$\uparrow$} & \textbf{XCOMET$\uparrow$} & \textbf{MetricX$\downarrow$} & \textbf{BLEU$\uparrow$} & \textbf{XCOMET$\uparrow$} & \textbf{MetricX$\downarrow$} \\
\midrule
\multicolumn{8}{r}{\textbf{High Resource (55.4M and 52.0M training data)}} \\
\midrule
MAP& 5 & -- &24.41& 81.26&5.18&19.34&73.56&7.57\\
 MBR& 5 & 5 & 25.85&82.44&4.19&16.88&74.14&7.17\\
\hdashline
MAP & 400 & -- &13.62&76.83&7.97&22.01\textsuperscript{††}&77.65&6.58 \\
QE reranking & 400 & 1 &\textbf{28.64}& 86.12&3.12&18.94&83.12&4.57\\
MBR & 400 & 400 & 28.23&86.20&3.17&19.48&80.09&5.43\\
 MBR& 400 & 17 &26.72&85.18&3.84&\textbf{23.05}\textsuperscript{††}&81.75&5.10\\
sMBR-PP & 400 & 17 &27.55&\textbf{86.47}\textsuperscript{†}& \textbf{3.00}\textsuperscript{†}&18.89&\textbf{83.41}\textsuperscript{†}&\textbf{4.47}\textsuperscript{†}\\
sMBR-BT & 400 & 17 &12.94&78.26&7.53&17.36&79.69&5.81\\
\bottomrule
\end{tabular}}
\caption{Compares sMBR with other decision rules for En$\rightarrow$De and En$\rightarrow$Ru in the \textbf{classic setup} with \textbf{epsilon sampling}. $|\mathcal{C}|$ and $|\mathcal{S}|$ indicate the number of candidate hypotheses and supportive hypotheses, respectively. For sMBR, we used $|\mathcal{S}|=$ 17 ``support hypotheses'' (1 original source + 16 quasi-sources). We performed paired bootstrap resampling over all methods against QE reranking; † and †† indicate significantly better than QE reranking within groups ($p < 0.05$ and $p < 0.01$, respectively; Multiple testing correction is not applied). The best in each group is marked in bold.}
\label{tab:smbr_eps}
\end{table*}

\section{Details of the classic setup}
\label{sec:classic_setup}
For the high-resource setup, training data consists of 27.7M and 26.0M parallel sentences for En$\rightarrow$De and En$\rightarrow$Ru, respectively, augmented with an equal amount of back-translation sentences. 
We use a single model without ensembling or language model reranking to focus on the impact of the proposed methods. 
For the low resource setup, we train two base Transformer models using the News-Commentary dataset\footnote{\url{data.statmt.org/news-commentary/v18.1/}}.
\section{Details of the LLM setup}
\label{sec:tower}
We use the target language to prompt the model to perform zero-shot translation. 
We used the following prompts during inference: 

En$\rightarrow$De:
\begin{center}
\code{Übersetzen Sie den folgenden Text \\vom Englischen ins Deutsche.\textbackslash n\\Englischen:\textbackslash n\{source\_text\}\textbackslash nDeutsche:}
\end{center}

En$\rightarrow$Ru:
\selectlanguage{russian}
\begin{center}
\code{Переведите следующий текст с \\английского на русский.\textbackslash n Английский:\textbackslash n\{source\_text\}\textbackslash nРусский:}
\end{center}
\selectlanguage{english}

Zh$\rightarrow$En:
\begin{center}
\code{Translate the following text from \\Chinese to English.\textbackslash nChinese:\\ \textbackslash n\{source\_text\}\textbackslash nEnglish:}
\end{center}

We completed all decoding on a server with four NVIDIA H100s with bfloat16 precision. 
For sMBR-PP on Zh$\rightarrow$En, we trained a mT5-based (\citealp{Xue2020mT5AM}) model for the paraphrase generator. See Appendix~\ref{sec:smbr_details} for details.
\begin{table*}[!t]
\scalebox{0.92}{
\centering
\begin{tabular}{lrrllllllll}
\toprule
\textbf{Decoding method } & & & \multicolumn{3}{c}{\textbf{En$\rightarrow$De}}  & \multicolumn{3}{c}{\textbf{Zh$\rightarrow$En}} \\
\cmidrule(r){4-6} \cmidrule(l){7-10}
~ & $|\mathcal{C}|$ & $|\mathcal{S}|$ & \textbf{BLEU$\uparrow$} & \textbf{XCOMET$\uparrow$} & \textbf{MetricX$\downarrow$} & \textbf{BLEU$\uparrow$} & \textbf{XCOMET$\uparrow$} & \textbf{MetricX$\downarrow$} \\
\midrule
w/ beam search & \multicolumn{8}{c}{\textbf{\textit{TowerInstruct-13B}}} \\
\midrule
 MAP& 5 & -- & 39.75&87.11 & 2.47 &24.94 & 89.13 &2.13 \\
 MBR& 5 & 5 & 39.84&87.29 & 2.46 &25.00 &89.14& 2.13\\
\hdashline
MAP & 128 & -- & 40.04&87.07 & 2.44 &24.96 &89.12& 2.14 \\
QE reranking & 128 & 1 & 40.11&87.35 & 2.40 &24.96& 89.22 & 2.11 \\
MBR & 128 & 128 & 40.07& 87.29&2.41 &\textbf{25.00} & 89.17 & 2.13 \\
 MBR& 128 & 17 & 40.14& 87.19& 2.41&24.96 &89.13& 2.14 \\
sMBR-PP & 128 & 17 & 40.15 &\textbf{87.45}\textsuperscript{†}& \textbf{2.36}&24.93 &\textbf{89.28}\textsuperscript{†} &\textbf{2.10}  \\
sMBR-BT & 128 & 17 & \textbf{40.18}& 87.39&2.37 &24.90 &89.21 & 2.11\\
\midrule
w/ top-k sampling \\
\midrule
 MAP& 5 & -- & 28.12& 84.85 & 3.31 &19.94 &88.14 &2.42 \\
 MBR& 5 & 5 & 25.52& 86.12 &3.12 &18.52 & 88.82 &2.31 \\
\hdashline
MAP & 128 & -- & \textbf{32.13}\textsuperscript{††} &86.17 &3.30 &21.90\textsuperscript{††} &88.56 &2.22 \\
QE reranking & 128 & 1 & 27.56&88.09 & 2.73&19.07 &90.10 & 1.93 \\
MBR & 128& 128 & 28.28\textsuperscript{†}& 88.68\textsuperscript{††} &2.56\textsuperscript{†}  &21.19\textsuperscript{††} & 90.23 &1.94\\
 MBR& 128 & 17 & 30.66\textsuperscript{††}& 88.76\textsuperscript{††} & 2.59 &\textbf{22.40}\textsuperscript{††} & \textbf{90.47}\textsuperscript{††} & \textbf{1.87}\\
sMBR-PP &128 & 17 & 25.20& \textbf{89.05}\textsuperscript{††}& \textbf{2.47}\textsuperscript{††}&18.85 &90.19 &1.91\\
sMBR-BT &128& 17 & 27.13& 88.48 &2.65&18.87 &89.82&2.03 \\
\midrule
w/ ancestral sampling \\
\midrule
 MAP& 5 & -- & 26.29&83.20 & 4.00 &19.07 &87.21 & 2.62 \\
 MBR& 5 & 5 & 24.20& 84.67& 3.62 &17.97 & 87.79 & 2.52\\
\hdashline
MAP & 128 & -- & \textbf{30.81}\textsuperscript{††}& 85.76 & 3.42 &\textbf{21.86}\textsuperscript{††} &88.57 & 2.32 \\
QE reranking & 128 & 1 & 27.10&87.57 & 2.89 &18.67 & 89.69 &2.07\\
MBR & 128 & 128 & 26.88& 87.59 &\textbf{2.75} &20.65\textsuperscript{††}  & 89.65 & \textbf{2.03} \\
 MBR&128 & 17 & 28.64\textsuperscript{††}& 87.50 & 2.78 &21.72\textsuperscript{††} & \textbf{90.02}\textsuperscript{†} & 1.97 \\
sMBR-PP & 128 & 17 & 25.78 &\textbf{87.91}\textsuperscript{†}& 2.86 &18.29 &89.70 &2.08\\
sMBR-BT & 128 & 17 & 26.78& 87.79 &2.84 &18.40 &89.41& 2.13 \\
\midrule
w/ epsilon sampling \\
\midrule
 MAP& 5 & -- & 30.24&86.06 & 3.32 &20.73 &88.15 &2.41 \\
 MBR& 5 & 5 & 28.10& 87.30 &2.95 &19.74 &88.96 & 2.20 \\
\hdashline
MAP & 128& -- & \textbf{32.64}\textsuperscript{††}&86.43 &3.22&23.12\textsuperscript{††} &89.14 &2.20 \\
QE reranking & 128 & 1 & 29.40& 88.76 &2.56 &19.88 &90.64 & 1.89 \\
MBR & 128 & 128 & 29.84&89.19\textsuperscript{†} & 2.46\textsuperscript{††} &22.01\textsuperscript{††} & 90.39& 1.90 \\
 MBR& 128 & 17 & 31.93\textsuperscript{††}& 88.83 & 2.60 &\textbf{23.34}\textsuperscript{††}& 90.43 & 1.87 \\
sMBR-PP & 128 & 17 & 27.19 & \textbf{89.47}\textsuperscript{††}&  \textbf{2.44}\textsuperscript{††} &19.87 &\textbf{90.70}\textsuperscript{†} &  \textbf{1.87}\\
sMBR-BT &128 & 17 & 28.73& 89.04 &2.50 &19.77 & 90.38& 1.98 \\
\bottomrule
\end{tabular}}
\caption{Compares sMBR with other decision rules for En$\rightarrow$De and Zh$\rightarrow$En in the \textbf{LLM setup}. $|\mathcal{C}|$ and $|\mathcal{S}|$ indicate the number of candidate hypotheses and supportive hypotheses, respectively. For sMBR, we used $|\mathcal{S}|=$ 17 ``support hypotheses'' (1 original source + 16 quasi-sources). We performed paired bootstrap resampling over all methods against QE reranking; † and †† indicate significantly better than QE reranking within groups ($p < 0.05$ and $p < 0.01$, respectively; Multiple testing correction is not applied). The best in each group is marked in bold.}
\label{tab:smbr_tower}
\end{table*}

\begin{table*}[!t]
\centering
\scalebox{1.1}{
\begin{tabular}{lrrllllllll}
\toprule
\textbf{Decoding method } & & & \multicolumn{3}{c}{\textbf{En$\rightarrow$Ru}}  \\
\cmidrule(r){4-6} \cmidrule(l){7-10}
~ & $|\mathcal{C}|$ & $|\mathcal{S}|$ & \textbf{BLEU$\uparrow$} & \textbf{XCOMET$\uparrow$} & \textbf{MetricX$\downarrow$} \\
\midrule
w/ beam search & \multicolumn{4}{c}{\textbf{\textit{TowerInstruct-13B}}} \\
\midrule
 MAP& 5 & -- & 29.46&89.51 & 3.00 \\
 MBR& 5 & 5 & 29.33&89.61 & 2.96 \\
\hdashline
MAP & 128 & -- & 29.50\textsuperscript{†}&89.58 & 3.00 \\
QE reranking & 128 & 1 & 29.33&89.65 & 2.97 \\
MBR & 128 & 128 & 29.38&89.69 & 2.97 \\
 MBR& 128 & 17 & \textbf{29.52}\textsuperscript{††}&89.57 & 3.00 \\
sMBR-PP & 128 & 17 & 29.36&\textbf{89.77}\textsuperscript{†} &\textbf{2.96}\\\
sMBR-BT & 128 & 17 & 29.39&89.69 &\textbf{2.96} \\
\midrule
w/ top-k sampling \\
\midrule
 MAP& 5 & -- & 23.22&86.79 & 3.69 \\
 MBR& 5 & 5 & 21.23&87.02 & 3.53 \\
\hdashline
MAP & 128 & -- & \textbf{28.07}\textsuperscript{††}&89.30 & 2.97 \\
QE reranking & 128 & 1 & 22.02&91.37 & 2.45\\
MBR & 128& 128 & 24.24\textsuperscript{††}&91.26 & 2.48\\
 MBR& 128 & 17 & 27.47\textsuperscript{††}&91.36 & 2.41\\
sMBR-PP &128 & 17 & 20.24&\textbf{91.49}\textsuperscript{††} &\textbf{2.40}\\
sMBR-BT &128& 17 & 21.27&90.80 & 2.51\\
\midrule
w/ ancestral sampling \\
\midrule
 MAP& 5 & -- & 23.28&86.79 & 3.69\\
 MBR& 5 & 5 & 21.12&87.18 & 3.51\\
\hdashline
MAP & 128 & -- & \textbf{28.32}\textsuperscript{††}&89.27 & 3.01 \\
QE reranking & 128 & 1 & 21.35&90.96 & 2.55\\
MBR & 128 & 128 & 23.77\textsuperscript{††}&90.98 & 2.53\\
 MBR&128 & 17 & 26.55\textsuperscript{††}&\textbf{91.23}\textsuperscript{††} & \textbf{2.47}\\
 sMBR-PP & 128 & 17 & 20.52&90.99 & 2.50\\
sMBR-BT & 128 & 17 & 21.63&90.78 & 2.57\\
\midrule
w/ epsilon sampling \\
\midrule
 MAP& 5 & -- & 26.19&88.53 & 3.20\\
 MBR& 5 & 5 & 23.92&89.56&2.87\\
\hdashline
MAP & 128 & -- &\textbf{29.61}\textsuperscript{††}&89.68&2.97 \\
QE reranking & 128 & 1 & 22.35&91.67 & 2.34\\
MBR & 128 & 128 & 25.90\textsuperscript{††}&91.33 & 2.35\\
MBR&128 & 17 & 28.94\textsuperscript{††}&91.54 & 2.40\\
sMBR-PP & 128 & 17 & 21.26&\textbf{92.12}\textsuperscript{††} &\textbf{2.20}\textsuperscript{††}\\
sMBR-BT & 128 & 17 & 22.47&91.68 &2.38 \\
\bottomrule
\end{tabular}}
\caption{Compares sMBR with other decision rules for \textbf{En$\rightarrow$Ru}, the NMT model is \textbf{\code{TowerInstruct-13B}}. $|\mathcal{C}|$ and $|\mathcal{S}|$ indicate the number of candidate hypotheses and supportive hypotheses, respectively. For sMBR, we used $|\mathcal{S}|=$ 17 support hypotheses (1 original source + 16 quasi-sources). We performed paired bootstrap resampling over all methods against QE reranking; † and †† indicate significantly better than QE reranking within groups ($p < 0.05$ and $p < 0.01$, respectively; Multiple testing correction is not applied). The best in each group is marked in bold.}
\label{tab:smbr_tower_ru}
\end{table*}

\section{Efficiency comparison}
\label{sec:speed_with_mbr}
In this section, we discuss the efficiency of sMBR-PP and compare it with QE reranking and MBR. For the latter, we compare the average decision time required for MBR, an optimized implementation of MBR (MBR-fast), and sMBR-PP to translate a single sentence. We ran each method five times on a single NVIDIA H100 with batch size 256 examples and then report the means.

As expected, the decision time required for sMBR-PP to translate a sentence is much larger than that of QE reranking. Specifically, the decision time of sMBR-PP consists of two parts: the time required to generate the quasi-source and the time required to calculate the quality-aware utility function. In fact, we find that the time required to generate the quasi-source is only a small part of the overall decision time, which is about 0.13 seconds for each sentence, while the large number of quality-aware utility functions requires 3.56 seconds. In contrast, the decision time for QE reranking is 0.21 seconds per sentence, which is much faster than sMBR-PP. 

Obviously, compared to QE reranking, sMBR-PP uses the quality perception utility function times ${(}$number of quasi-sources ${+1)}$.

Next, we compare the decision time of sMBR-PP with that of the standard MBR. 
In MBR, the COMET model can be decomposed into a sentence encoder ${f_{emb}}$ for computing sentence embeddings, and a simple estimator ${f_{est}(\cdot,\cdot,\cdot)}$ based on a multilayer perceptron. For a source ${x}$, a candidate hypothesis ${h}$, and support hypotheses ${h_s \in \mathcal{S}}$, COMET-based MBR first computes the source embedding ${x^{emb}}$, the candidate hypothesis embedding ${h^{emb}}$, and a set of support hypotheses embeddings $\mathcal{S}^{emb}$ , using ${f_{emb}}$. Then, the MBR score of ${h}$ $score^{\mathit{MBR}}_{h}$ can then be computed as:
\begin{align}
score^{\mathit{MBR}}_{h}
= \frac{1}{|\mathcal{S}|} \sum_{h_s^{emb} \in \mathcal{S}^{emb}} {f_{est}(x^{emb},{h_s}^{emb},h^{emb})}.
\end{align}
When ${\mathcal{S} = \mathcal{C}}$, the cost of computing utility for all candidate hypotheses in a naive MBR implementation is ${\mathcal{O}(|\mathcal{C}|^2)}$, implying a quadratic cost for both ${f_{emb}}$ and ${f_{est}(\cdot,\cdot,\cdot)}$.

However, MBR-fast optimizes embedding computation by recognizing that the embedding any sentence in a triple $(x,h,h_s)$ is independent of the other elements. 
By pre-computing sentence embeddings independently for all sources and hypotheses, MBR-fast avoids duplicate ${f_{emb}}$ computations and reduces its cost to ${\mathcal{O}(|\mathcal{C}|)}$ when ${\mathcal{S} = \mathcal{C}}$.
The estimator ${f_{est}(\cdot,\cdot,\cdot)}$ still has a quadratic cost ${\mathcal{O}(|\mathcal{C}|^2)}$ since the order of elements within the triple affects the output,
but it is computationally cheaper compared to a ${f_{emb}}$ consisting of multiple transformer blocks. Note that this optimization is not universal due to the fact that it takes advantage of the particular architecture of COMET.

In contrast, the COMET-QE model used in sMBR consists of an encoder ${f^{QE}_{emb}}$ that takes the concatenated source and hypothesis as input and outputs their joint embedding, and an estimator ${f^{QE}_{est}}$. The joint embeddings must be computed separately for each source-hypothesis pair, resulting in a cost of ${\mathcal{O}(K \times |\mathcal{C}|)}$ for both ${f^{QE}_{emb}}$ and ${f^{QE}_{est}}$.

Table~\ref{tab:speed} shows the measurement results. sMBR is faster than the naive implementation of MBR because it uses a smaller number of support hypotheses. However, it is much slower than the optimized implementation of MBR due to the difficulty of further optimizing its utility function itself.

In summary, while sMBR-PP significantly improves translation quality compared to QE reranking and has competitive performance to MBR, there is still room for improving its efficiency to match optimized COMET-based MBR decoding.
\begin{table}[H]
\scalebox{0.85}{
\centering
\begin{tabular}{lcccc}
\toprule
~ & MBR & MBR-fast & sMBR-PP \\
\midrule
Decision time & 135.29 s & 0.32 s & 3.56 ( + 0.13) s \\
\bottomrule
\end{tabular}}
\caption{Decision time for translating a sentence: measured on \code{newstest2020} in En$\rightarrow$De. For sMBR-PP, the number in parentheses is the quasi-source generation time. The batch size is 256.}
\label{tab:speed}
\end{table}
\section{Impact of the number of candidate hypotheses}
\label{sec:candidate}
\begin{figure*}[!htp]
    \centering
    \includegraphics[width=0.985\linewidth]{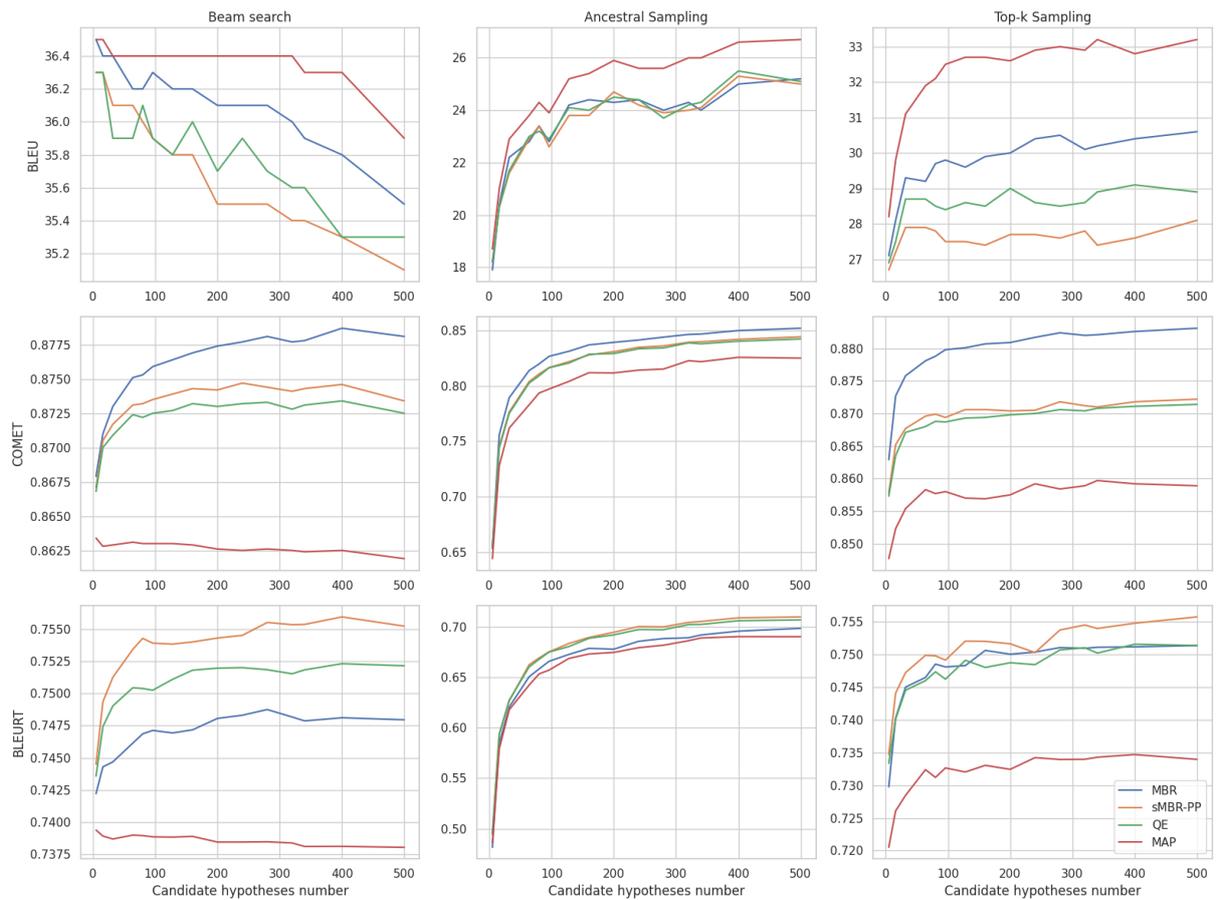}
    \caption{Impact of the number of candidate hypotheses on the evaluation metrics: in the En$\rightarrow$De high resource setup. The horizontal axis indicates the number of candidate hypotheses and the vertical axis indicates the evaluation indicators.}
    \label{fig:hypo_num}
\end{figure*}

We explored the impact of the number of candidate hypotheses on the evaluation metrics in an En$\rightarrow$De high resource setting. Figure~ \ref{fig:hypo_num} shows the results. We find that 400 is an appropriate number, as more candidate hypotheses bring small performance gains and lead to higher costs.

\section{Using LLM to generate quasi-sources for sMBR-PP}
\label{sec:smbr_gpt}
We show in \label{sec:anal_syn} the results of investigating the nature of quasi-sources in sMBR-PP and sMBR-BT based on Self-BLEU and semantic similarity. 
These results suggest that quasi-sources with richer surface forms and greater semantic similarity to the original source may lead to better translation quality.
In fact, during the early stages of this research, we tried using \code{GPT4-0125} (\citealp{Achiam2023GPT4TR}), the state-of-the-art LLM at the time, to generate quasi-sources for sMBR-PP. 

\begin{table}[h]
\centering
\scalebox{0.93}{
\begin{tabular}{lccc}
\toprule
~ & \textbf{XCOMET$\uparrow$} & \textbf{MetricX$\downarrow$} \\
\midrule
sMBR-PP (T5) & 86.52 &  4.14\\
sMBR-PP (GPT4) & 87.10  & 4.09 \\
\bottomrule
\end{tabular}}
\caption{Comparison of sMBR-PP performance using different paraphrase generators: Experiments conducted on high-resource sub-setup, En$\rightarrow$Ru language pair, using beam search to generate candidate hypotheses.}
\label{tab:gptvst5_en2ru}
\end{table}
Table~\ref{tab:gptvst5_en2ru} shows our results on the classic setup, En$\rightarrow$Ru language pair, high-resource sub-setup, using beam search to generate candidate hypotheses.

We found that sMBR-PP based on \code{GPT4-0125} achieved better performance on both XCOMET and MetricX. 

\begin{table}[h]
\centering
\scalebox{0.82}{
\begin{tabular}{lccc}
\toprule
~ & Self-BLEU$\downarrow$ & Semantic Similarity$\uparrow$ \\
\midrule
sMBR-PP (T5) & 45.95 &  92.83\\
sMBR-PP (GPT4) & 18.67 &  93.46\\
\bottomrule
\end{tabular}}
\caption{Analyzing of quasi-sources: analyzed on the En$\rightarrow$Ru \code{generaltest2023}, high resource. Lower Self-BLEU means richer surface diversity; higher semantic similarity means closer semantics to the original source.}
\label{tab:synt_src_en2ru}
\end{table}
We investigated the properties of quasi-sources generated by \code{GPT4-0125} using the same method as in \label{sec:anal_syn}, and the results are presented in Table~\ref{tab:synt_src_en2ru}.

We observed that quasi-sources generated by \code{GPT4-0125} had lower Self-BLEU than those generated by \code{T5}, while maintaining similar semantic similarity. This indicates that quasi-sources generated by \code{GPT4-0125} have richer surface forms.

Therefore, 
we believe that using paraphrases with richer surface forms as quasi-sources can indeed improve the performance of sMBR-PP. 
However, considering that research based on non-open models like \code{GPT4-0125} would make our work difficult to reproduce—after all, \code{GPT4-0125} produces different outputs for identical inputs even with a specified random seed or temperature of 0, and \code{GPT4-0125} may not be accessible in the future.
On the other hand, although sMBR-PP based on \code{T5} doesn't perform as well on XCOMET and MetricX as sMBR-PP based on \code{GPT4-0125}, 
\code{T5} is an open model that allows us to ensure the reproducibility of our research and support our conclusions. 
Therefore, we chose T5-based sMBR-PP as our main experimental setup.
Nevertheless, recent research on open LLMs has made significant progress, and we will explore using open LLMs as alternative experimental setups for sMBR-PP in future work.

\section{Generating quasi-sources for sMBR-PP with diverse beam search}
\label{sec:dbs}
The analysis results in \label{sec:anal_syn} suggest that the poor performance of sMBR-BT may be due to the lack of diversity in the surface form of the quasi-source. 
This could be because we used simple beam search to generate the quasi-source.

Diverse Beam Search (DBS) (\citealp{dbs}) is an improved beam search method that can generate more diverse text. 
We tried using different beam search methods to generate quasi-sources for sMBR-BT in the classic setup.

\begin{table}[h]
\centering
\scalebox{0.93}{
\begin{tabular}{lccc}
\toprule
~ & \textbf{XCOMET$\uparrow$} & \textbf{MetricX$\downarrow$} \\
\midrule
QE ranking &86.48 &3.22 \\
sMBR-BT (BS) & 86.17 &  3.33\\
sMBR-BT (DBS) & 86.19  & 3.29 \\
\bottomrule
\end{tabular}}
\caption{Comparison of sMBR-BT performance based on different search methods: Experiments conducted on high-resource sub-setup, En$\rightarrow$De language pair, using beam search to generate candidate hypotheses.}
\label{tab:dbs}
\end{table}
Table~\ref{tab:dbs} shows the experimental results for the En$\rightarrow$De task (classic setting, high-resource subset, using beam search to generate candidate hypotheses), where sMBR-BT (BS) indicates using simple beam search to generate the quasi-source, while sMBR-BT (DBS) indicates using DBS. 
We found that using DBS to generate quasi-sources only slightly improved the performance of sMBR-BT, but it was still worse than QE reranking.

We will attempt to use more improved methods (such as sampling-based generation methods) to generate quasi-sources to enhance the performance of sMBR-BT as future work.
\label{sec:smbr_gpt}
\end{document}